\documentclass[11pt,twoside]{article}
\usepackage{epsfig}
\usepackage{graphics}
\usepackage{graphicx}
\usepackage{natbib,amssymb,lineno}

\title{Towards 3D Visualization of Video from Frames}
\author{Slimane Larabi\\
 RIIMA Laboratory, Computer Science Department\\
 USTHB University, 16111 Algiers, Algeria\\
 email: slarabi@usthb.dz
 }
\date{}

\chardef\bslchar=`\\ 

\providecommand{\qedsymbol}{\leavevmode
  \hbox to.77778em{%
  \hfil\vrule
  \vbox to.675em{\hrule width.6em\vfil\hrule}%
  \vrule\hfil}}

\catcode`\|=0
\begingroup \catcode`\>=13 
\gdef\?#1>{{\normalfont$\langle$\textit{#1}$\rangle$}}
\gdef\0{\relax}
\endgroup
\def\<#1>{{\normalfont$\langle$\textit{#1}$\rangle$}}

\def\latex/{{\protect\LaTeX}}

\usepackage{url}
\usepackage[breaklinks]{hyperref}

\begin{document}
\maketitle
\markboth{Towards 3D Visualization of Video from Frames}
{Towards 3D Visualization of Video from Frames}

\begingroup
\small
\tableofcontents
\endgroup


\newpage 
\begin{abstract}
  We explain theoretically how to reconstruct the 3D scene from successive frames in order to see the video in 3D.
To do this, features, associated to moving rigid objects in 3D, are extracted in frames and matched. The vanishing point computed in frame corresponding to the direction of moving object is used for 3D positioning of the 3D structure of the moving object.
First experiments are conducted and the obtained results are shown and publicly available. They demonstrate the feasibility of our method.
We conclude this paper by future works in order to improve this method tacking into account non-rigid objects and the case of moving camera.
\end{abstract}
\section{Introduction}

Our aim is to reconstruct the 3D scene from a video sequence of moving object with a static camera.
Figure \ref{fig0} shows a sample of four frames taken from a video. The proposed scene reconstruction is based on the temporal information in the image plane due to the moving object in the scene and on vanishing points inferred from displacement of keypoints of rigid object in the scene.
We give in section \ref{sec1} the basic principle of our method. The reconstruction of all keypoints of successive frames in explained in section \ref{algo}. The obtained results with synthetic data are presented and commented in section \ref{exp}. We conclude this paper with some future works in section \ref{concl}.

\begin{figure}[htb]
\centering
\includegraphics[width=3cm]{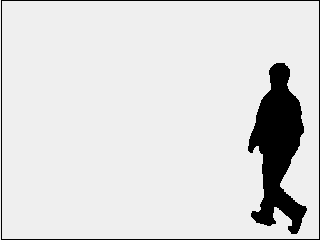}
 \includegraphics[width=3cm]{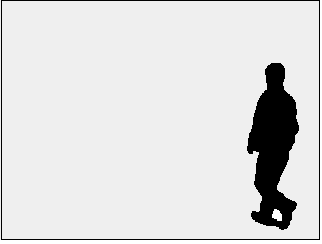}
 \includegraphics[width=3cm]{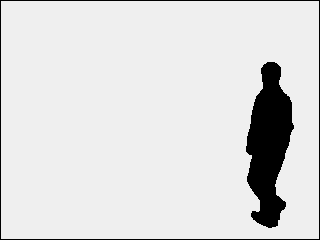}
 \includegraphics[width=3cm]{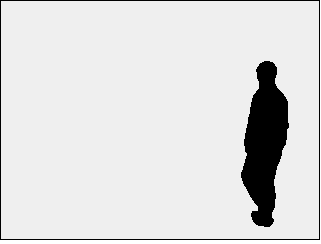}\\
\includegraphics[width=6cm]{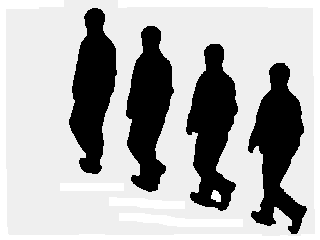}
\caption{A set of four frames and the aimed reconstructed scene.}
\label{fig0}
\end{figure}

\section{Basic Principle of our method}\label{sec1}

\subsection{Vanishing points from joining matched keypoints in successive frames}

Let $(a',c')$ be the images of two key points of a moving object in the $3D$ scene located on the frame $t$. $(a,c)$ are the corresponding 3D points of (a',c') (see figure \ref{fig0}).
Let $(b',d')$ be the images of two key points of a moving object in the $3D$ scene located on the frame $t+\delta t$. $(b,d)$ are the corresponding 3D points of $(b',d')$ such that $(b',d')$ are the matches of $(a',c')$ in the sense that $(a', b')$ are the images of the same moving 3D point and $(c', d')$ are the images of the same moving 3D point (see \ref{fig0}.
If we assume that the 3D object is rigid, then $(ab)// (cd)$, then the lines $(a'b')$ and $(c'd')$ intersect in the vanishing point $\omega$ associated to the direction of $(ab)$ (see figure \ref{fig0}).

\begin{figure}[htb]
\centering
\includegraphics[width=10cm]{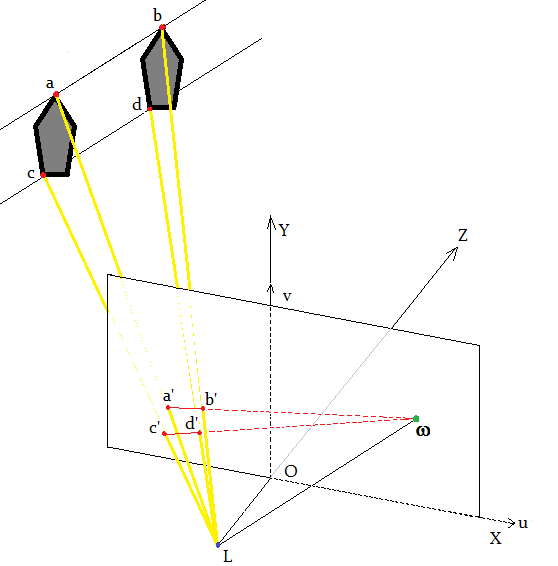}
\caption{The vanishing point $\omega$ located as the intersection of images of parallel 3D lines $(ab)$ and $(cd)$. }
\label{fig0}
\end{figure}

\subsection{$(X,Z)-$Positioning of key points from two frames}

Without loss of generality, we explain our method using the deux dimensions of the $3D$ space (the $Y$ component is ignored for this explanation). This is equivalent to see the drawn figure with a top-view.
The image plane is then represented by the $Ou$ axis and each $3D$ point is represented by its $X$ and $Z-$coordinates.\\

\begin{figure}[htb]
\centering
\includegraphics[width=10cm]{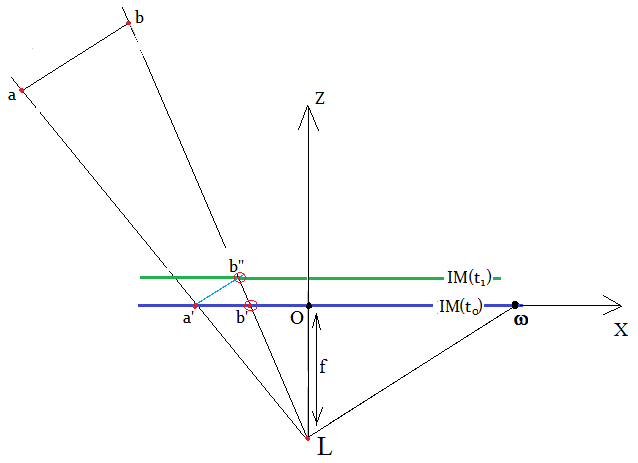}
\caption{Top view of the scene. $(X,Z)-$Positioning of key points from two frames. Note that $b''$ is the same point as $b'$. To obtain this, two translations are performed $(d_X, d_Z)$.}
\label{fig1}
\end{figure}

Figure \ref{fig1} illustrates the reconstruction process where $OXYZ$ is 3D referential such that the $O$ is principal point, $OX$ axis coincides with rows, $OY$ axis coincides with columns and $OZ$ coincides with the optical axis $LO$ (L is the projection center).\\

We consider two frames $t_0$ and $t_1$ reflecting the motion in 3D. The key point $a'$ in $t_0$ moves to $b'$ in the frame $t_1$.
Let $b''$ be the intersection point of the projection ray $Lb'$ with the line passing by $a'$ and parallel to $(L\omega)$.
If we place the image plane  $IM{t_1}$ parallel to $IM(t_0)$ such that $b''$ belongs to $IM^{t_1}$ and $L, b', b''$ are aligned (see figure \ref{fig1}), then the segment $ab$ is reconstructed as $a'b''$ because it is parallel to $ab$ with a ratio $r=La/La'$. This ration will be used in the next for the rest of reconstruction.

\subsection{$Y-$Positioning of key points from two frames}

\begin{figure}[htb]
\centering
\includegraphics[width=10cm]{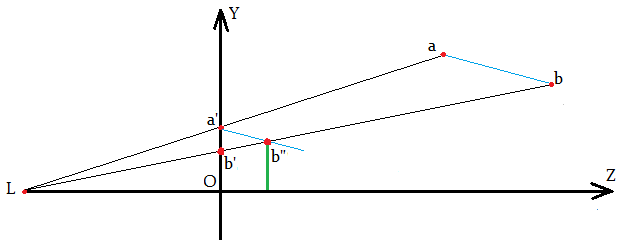}
\caption{$(Y-Z)$ view of the scene. Positioning of the plane $IM(t_1)$ such that the points $L, b", b"$ aligned and $a'b"$ parallel to $L\omega$.}
\label{fig2}
\end{figure}

To locate the point $b"$, we apply the analytic geometry. Knowing that $L, b', b''$ aligned and using the assumed data: focal length $f$ (equal to $LO$), positions of $O, a', b'$ in the frame $IM(t_0)$, we can compute the Y-coordinate of $b"$ by the equation \ref{eq1} (see figure \ref{fig2}).
\begin{equation}\label{eq1}
  Y(b")/Y(b')=(f+d_Z)/f
\end{equation}

\subsection{Reconstruction of the keypoints of the following frames}

\begin{figure}[htb]
\centering
\includegraphics[width=7cm]{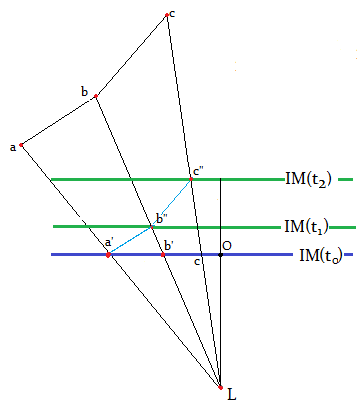}
\caption{Positioning of the third $IM(t_2)$ in order to reconstruct the point $c"$}
\label{fig3}
\end{figure}

If the point $b$ moves towards $c$ in the third frame, the same principle is applied considering the start point $b"$
instead of $a'$ and using the vanishing point $\omega$ associated to the direction of $bc$. A new plane $IM^{t_2}$ is created to which $c"$ belongs, the segment $bc$ is reconstructed as $b"c"$ (see figure \ref{fig3}).

\section{Reconstruction of all keypoints of successive frames}\label{algo}

\subsection{Summary}

Each keypoint of the second frame must be located in the 3D space. The explained method applied to keypoints associated to planar 3D points of the same object will produce 3D points appertaining to the first positioned plane $IM(t_0)$. To each other keypoint, a 3D point is associated appertaining to a different plane parallel to $IM(t_0)$.

\subsection{Algorithm}

\textbf{Begin}\\
-   \textbf{Input}: $F_{i}, i=0..k$ are the frames of the video sequence.\\
-   \textbf{Output}: 3D position of points corresponding to each keypoint in the frame $F_i$\\
-   \textbf{FOR}{Each $F_{i}$, $(i = 0..k-1)$}\\
-   \textbf{DO}\\
-   \textbf{FOR}{Each keypoint $p^i_j$ of the frame $F_{i}$}\\
-   \textbf{DO}\\
-    Locate the associated keypoint $p^{i+1}_{l}$ in the frame $F_{i+1}$\\
-    Estimate the direction of $p^i_j p^{i+1}_{l}$\\
-    Compute the 3D position of the associated keypoint $p^{i+1}_{l}$ of the frame $F_{i+1}$\\
- \textbf{ENDFOR}\\
- \textbf{ENDFOR}

\section{Experiments}\label{exp}

\begin{figure}[htb]
\centering
\includegraphics[width=6cm]{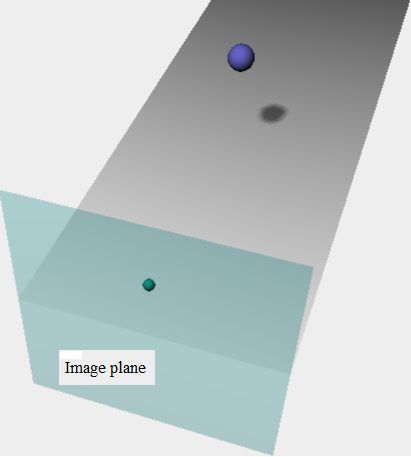}
\includegraphics[width=6cm]{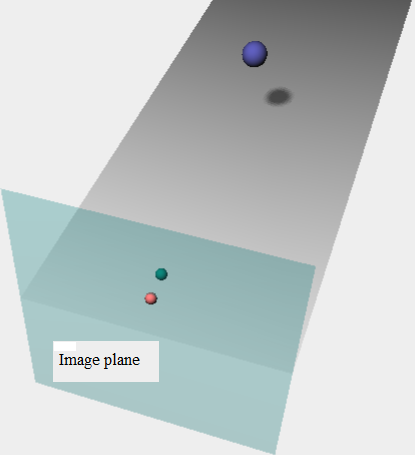}\\
\includegraphics[width=6cm]{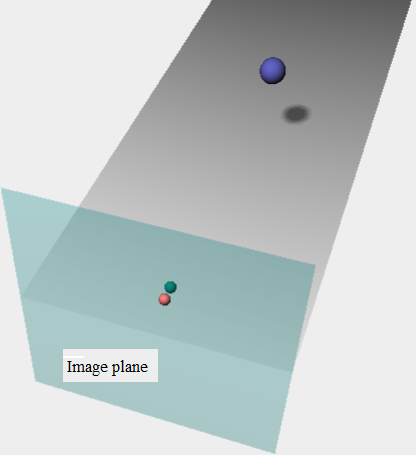}
\includegraphics[width=6cm]{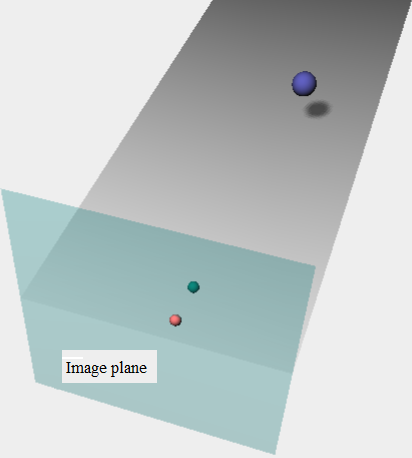}
\caption{Reconstruction of moving spheres (in blue color) from their images (red color). The first reconstructed sphere takes the position of its image. The green spheres are the reconstructed ones}
\label{fig6}
\end{figure}

We give an example of 3D reconstruction of a scene where the great sphere(with blue color in figure \ref{fig6}) moves from (0,10,20)to (0,8,24), (2,8,22), (4,5,26). Figure \ref{fig6} illustrates the reconstructed sphere (small with red color) using only from their images shown in small size in blue color The first reconstructed sphere is identical to the image (projection of the great sphere) on the first frame. The shadow presents on the gray plane helps the understanding of the 3D positioning.

\begin{figure}[htb]
\centering
\includegraphics[width=10cm]{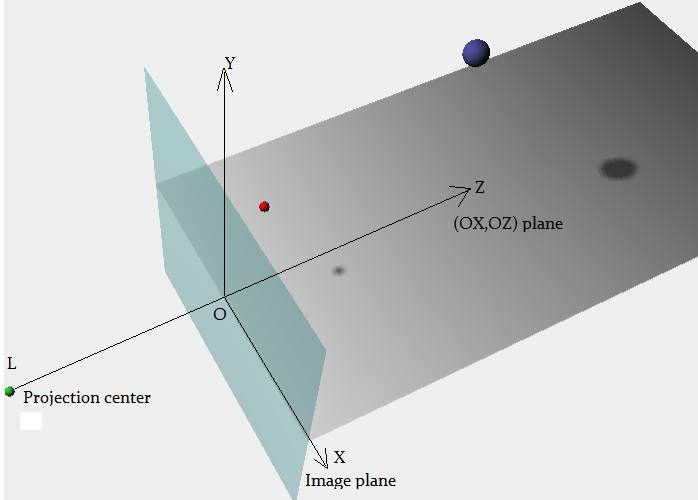}
\caption{First screenshot of the 3D reconstruction of rotating sphere.}
\label{figX1}
\end{figure}

\begin{figure}[htb]
\centering
\includegraphics[width=10cm]{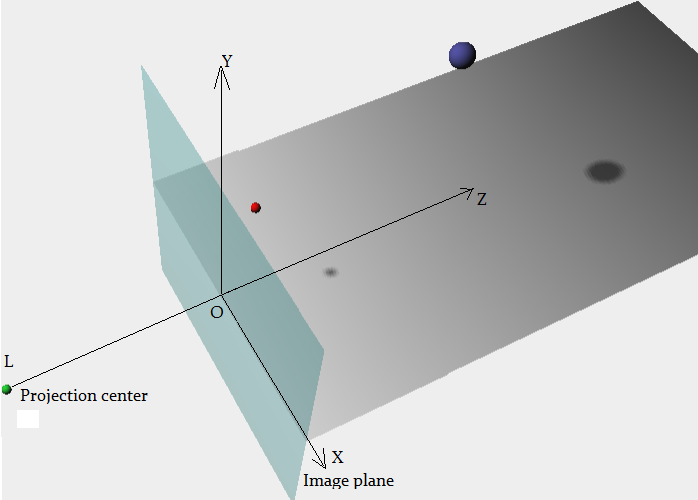}
\caption{Second screenshot of the 3D reconstruction of rotating sphere.}
\label{figX2}
\end{figure}

\begin{figure}[htb]
\centering
\includegraphics[width=10cm]{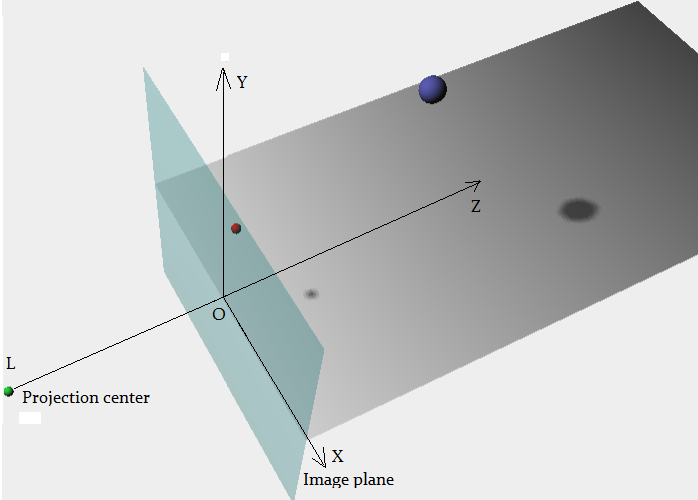}
\caption{Third screenshot of the 3D reconstruction of rotating sphere.}
\label{figX3}
\end{figure}

\begin{figure}[htb]
\centering
\includegraphics[width=10cm]{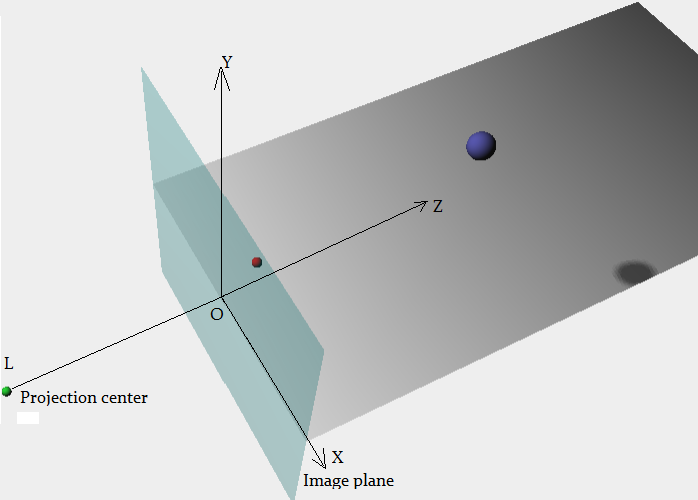}
\caption{Fourth screenshot of the 3D reconstruction of rotating sphere.}
\label{figX4}
\end{figure}

\begin{figure}[htb]
\centering
\includegraphics[width=10cm]{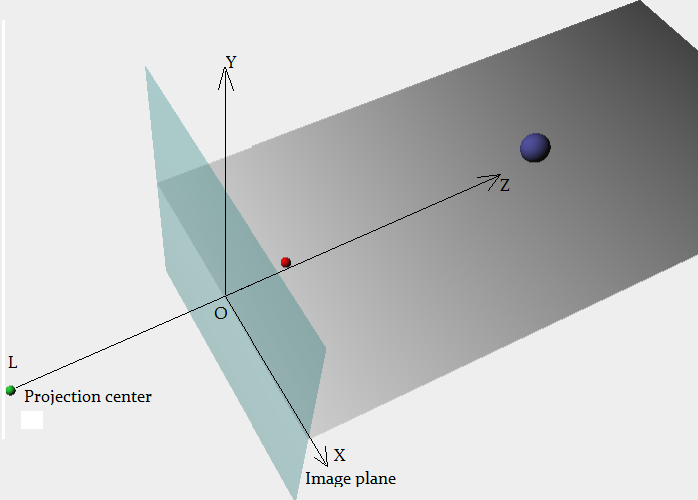}
\caption{Fifth screenshot of the 3D reconstruction of rotating sphere.}
\label{figX5}
\end{figure}

\begin{figure}[htb]
\centering
\includegraphics[width=10cm]{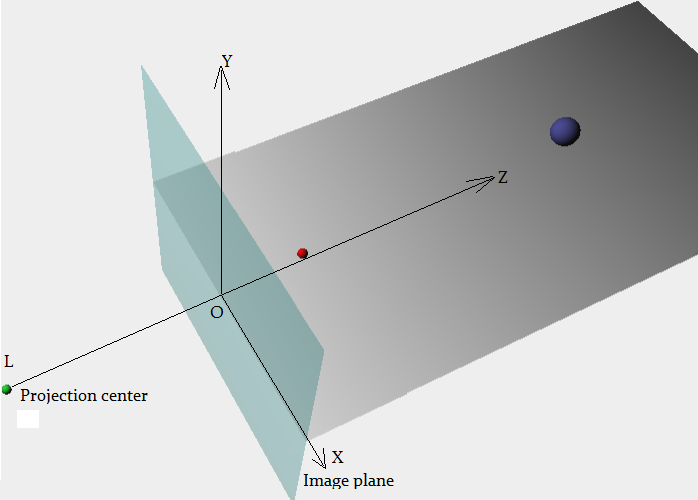}
\caption{Sixth screenshot of the 3D reconstruction of rotating sphere.}
\label{figX6}
\end{figure}

Figures \ref{figX1}$-$\ref{figX6} shows a second experiment with a rotating of the 3D object (great sphere) without visualizing position of its images on the image plane. Reconstructed spheres, using only their images, and the direction of moving sphere, are represented in red color.
Note that, here, we have not implemented the key features location and tracking for the computation of 3D points directions. Our aim in this paper is to give the mathematical reasoning and to show the feasibility of the proposed solution.
The results of this experiment is publicly available in \citep{myref}.

\section{Conclusion and Future Works}\label{concl}


In this paper we presented a new method to reconstruct the scene from video sequence. Our goal is to develop a codec that take as input a video sequence (mp4) and visualize the moving object in 3D. 
Our future works will be devoted for the improvement of this method by:\\
- Tacking into account non-rigid objects, the problem posed here is how to determine vanishing points.\\
- In the case of moving camera. In this case, the geometry reasoning must be modified.



\end{document}